\documentclass{article}

     \PassOptionsToPackage{numbers, compress, square, comma}{natbib}

\usepackage[final]{neurips_2020_ml4ad}

\usepackage[utf8]{inputenc} 
\usepackage[T1]{fontenc}    
\usepackage{hyperref}       
\usepackage{url}            
\usepackage{booktabs}       
\usepackage{amsfonts}       
\usepackage{nicefrac}       
\usepackage{microtype}      
\usepackage{amsmath,graphicx}

\title{An Evaluation of RGB and LiDAR Fusion for Semantic Segmentation}

\author{%
  Amr S. Mohamed \\
  The American University in Cairo\\
  \texttt{amrsaeed@aucegypt.edu} \\
  \And
  Ali Abdelkader\\
  The American University in Cairo\\
  \texttt{aly.abdelkader@aucegypt.edu}
  \And
  Mohamed Anany\\
  The American University in Cairo\\
  \texttt{moerafaat@aucegypt.edu}
  \And
  Omar El-Behady\\
  The American University in Cairo\\
  \texttt{donomar@aucegypt.edu}
  \And
  Muhammad Faisal\\
  The American University in Cairo\\
  \texttt{mfaisal@aucegypt.edu}
  \And
  Asser Hangal\\
  The American University in Cairo\\
  \texttt{asser.hangal@aucegypt.edu}
  \And
  Hesham M. Eraqi\\
  The American University in Cairo\\
  \texttt{heraqi@aucegypt.edu}
  \And
  Mohamed N. Moustafa\\
  The American University in Cairo\\
  \texttt{m.moustafa@aucegypt.edu}
}

\begin{document}

\maketitle

\begin{abstract}
LiDARs and cameras are the two main sensors that are planned to be included in many announced autonomous vehicles prototypes. Each of the two provides a unique form of data from a different perspective to the surrounding environment. In this paper, we explore and attempt to answer the question:\textit{ Is there an added benefit by fusing those two forms of data for the purpose of semantic segmentation within the context of autonomous driving?} We also attempt to show at which level does said fusion prove to be the most useful. We evaluated our algorithms on the publicly available SemanticKITTI dataset. All fusion models show improvements over the base model, with the mid-level fusion showing the highest improvement of \textbf{2.7\%} in terms of  mean Intersection over Union (mIoU) metric.
\end{abstract}

\section{Introduction}
\label{sec_one}
One important task in the field of autonomous driving is semantic segmentation which involves classifying objects to a set of predefined labels to provide applications with an understanding of the environment. There are different techniques and they can be broadly classified to belong to one of the following categories: semantic segmentation on camera images \((RGB)\), semantic segmentation on LiDAR point clouds \((XYZD)\), and a combination of the two.

In this paper, we explore the added value of applying fusion on \(RGB\) and \(XYZD\) data within the context of semantic segmentation. In addition to that, we explore the effect of different levels of fusion on the task of semantic segmentation. We explore the three levels of fusion: early, mid, and late and show that, while the three levels of fusion improve over their individual counterparts, the mid-level fusion improves the most. We compare with LiDAR only based approaches.

The rest of the paper is organized as follows. Section~\ref{section:sec_two} presents the related work. Section~\ref{section:sec_three} introduces the proposed architectures. Section~\ref{section:sec_four} presents the dataset, experiments, and results. Section~\ref{section:sec_five} presents the conclusion of our research. Finally, Section~\ref{section:sec_six} presents the future work to be done.

\section{Related Work}
\label{section:sec_two}
\subsection{2D Semantic Segmentation}
The task of semantic segmentation on images has been extensively studied and can be considered a mature field. Previous work has utilized multiple approaches to the task through the utilization of Convolutional-Neural-Networks (CNNs).
Fully Convolutional Neural Networks are able to take the whole image as an input and perform fast and accurate inferences on it, making them appealing for semantic segmentation. However, a major drawback is the predefined fixed size of their receptive field, which may lead to object fragmentation and mislabeling if the objects have a different size than the receptive field.
Different approaches have arose in the past decade, namely DeepLab \cite{deeplabv3}, currently considered the state of the art in 2D Image Semantic Segmentation.   

\subsection{3D Semantic Segmentation}
Semantic segmentation on 3D point clouds has been approached in different ways of terms of representation. Some methods use the raw input data in the form of point clouds like \cite{PointCNN, PointSIFT}, others use a voxelized 3D space \cite{segcloud, multiscale_voxel}, and others utilize a spherical projection of the data as in \cite{squeezesegv2}. 

An example of work that uses raw input data is PointNet \cite{pointnet}. This classification network takes \textit{n} points as input, applies input and feature transformations, and then aggregates point features by max pooling. The output is classification scores for \textit{k} classes. The segmentation network is an extension to the classification net. It concatenates global and local features and outputs per point scores.
Evaluation was performed using Stanford 3D semantic dataset \cite{s3dis}. The dataset is scanned in 6 areas including 271 rooms. PointNet achieved an mIoU of 47.71\%

Other point cloud segmentation techniques use a volumetric representation of the point cloud, these methods divide the point cloud into voxels (the equivalent of a pixel in 3D space). An example of such method is \cite{segcloud}. The model was tested on four data sets; Semantic3D \cite{semantic3d}, the Large-Scale 3D Indoor Spaces Dataset S3DIS \cite{s3dis}, NYUV2 \cite{nyuv2}, and KITTI \cite{KITTI}. The RGB features were used on all datasets with the exception of KITTIT. The model achieved an mIoU of 61.30\%, 48.92\%, 43.45\%, 36.78\% on the datasets respectively. However, such methods require high computational power, encouraging researchers to resort to a fusion of 3D data and 2D data.

Finally, one of the most famous architectures that work on the spherical projection of the point cloud is SqueezeSeg \cite{squeezeseg, squeezesegv2}. First, a spherical projection is adopted to transform sparse, irregularly distributed 3D point clouds to dense 2D grid representations. This approach involves multiple steps. Firstly, the point cloud is projected into a sphere through an equation that is applied to each point. Only the front view area is considered and is divided into grids. Five features were used for each point namely the 3 Cartesian coordinates \(x, y, z\), remission (intensity), and range data. Secondly, the results of the projection is fed into the network which uses deconvolution modules to up-sample feature maps. Finally, a Conditional Random Field (CRF) is used to refine the label maps generated by the CNN. RangeNet++ \cite{milioto2019iros} follows a similar approach using a modified version of Darknet \cite{darknet53} with a GPU implementation of K-nearest neighbor instead of CRF.  

\subsection{Fusion}
Fusion based methods work by fusing data from different sensors to achieve better results. Fusion can be applied at different levels in the model. Mainly, early, mid, or late level fusion. Most methods incorporate an early fusion approach where the RGB data is complemented with the XYZ one. One model, Superpoint Graphs \cite{SPG}, represents 3D point clouds as a collection of interconnected simple shapes similar to the superpixel methods for image segmentation. The nodes of the graph represent the simple shapes, while the edges describe their adjacency relationship. This representation has multiple advantages: it considers entire objects as a whole instead of individual points or voxels, it is able to describe the relationship between adjacent objects, and the size of the SPG is defined by the number of simple shapes in it rather than by the number of points. While SPG outperforms many other models, it's not efficient on sparse data as a result of its pruning method and its runtime is higher than other comparable methods. 
In \cite{valeo_paper}, early and mid-level fusion architectures are proposed that take as input a polar grid representation of LiDAR data and camera. A downside of such approach is that it does not make full use of all camera pixels.  

\section{Proposed methods}
\label{section:sec_three}
In the following sections, we describe the architectures used for fusing inputs from LiDARs and Cameras to perform semantic segmentation with respect to multiple levels of fusion.

\subsection{Input data}
Instead of using raw point cloud, Polar Grid Map (PGM) is used as representation for LiDAR data. As shown in \cite{squeezeseg}, PGM representation is obtained by spherically projecting points and using their polar coordinates to position them inside a 2D grid. The result is a volume of shape \(H \times W \times C\) where \(H\) is the number of vertical channels and is determined by the LiDAR used to collect the data. For the SemanticKitti dataset \cite{SemnaticKitti}, the Velodyne HDL-64E LiDAR is used which has 64 vertical channels thus \(H=64\). However, while LiDAR's horizontal field of view covers 360 degrees, for the purposes of this work, only points lying in the field of view overlapping with camera frame are considered which limits our FoV to the front facing \(80^{\circ}\) horizontally and \(20^{\circ}\) vertically. To create the PGM, the horizontal \(80^{\circ}\) FoV was discretized into 512 polar grid cells so \(W=512\). \(C\) represents the number of features for each point. Our features are the 3 Cartesian coordinates (\(x, y, z\)), the 3 color values (\(r, g, b\)), the range, the intensity measurement, and 2 label channels (\(L1, L2\)). The color \(r, g, b\) values for each LiDAR point is obtained by projecting LiDAR point cloud on image frame using the relative calibration matrices. 

\subsection{Network architectures} Based on findings of \cite{SemnaticKitti}, SqueezeSegV2 \cite{squeezesegv2} was chosen as our backbone model because it provides the best trade off between inference time and performance. SqueezeSegV2's architecture is based on its predecessor SqueezeSeg \cite{squeezeseg}. It's input is the PGM representation described in Input Data Section which is a tensor of shape \(64 \times 512 \times C\). The model outputs a label map of shape \(H \times W \times 1\). For the purpose of this paper, layers up to fire 9 \cite{squeezesegv2} would be referred to as SqueezeSegV2 encoder while the rest of the model would be referred to as SqueezeSegV2 decoder.    

\subsubsection{Early-level Fusion}
Early-level Fusion can be defined as combining information from two modalities in the input space before providing it to a model. This leads to the question of whether to add the LiDAR information, range, and intensity to image pixels or to add the color information of image pixels to LiDAR PGM cells. Given that the number of LiDAR points in a scan is much lower than the number pixels in its corresponding image, the first approach would lead to a sparse representation of LiDAR information in pixel space. Therefore, the second approach was chosen in which SqueezeSegV2 would be used as an early fusion model after modifying its input tensor to have 8 channels: \( x, y, z, intensity, range, r, g ,b\)

\begin{figure}
  \centering
  \includegraphics[width=0.6\textwidth]{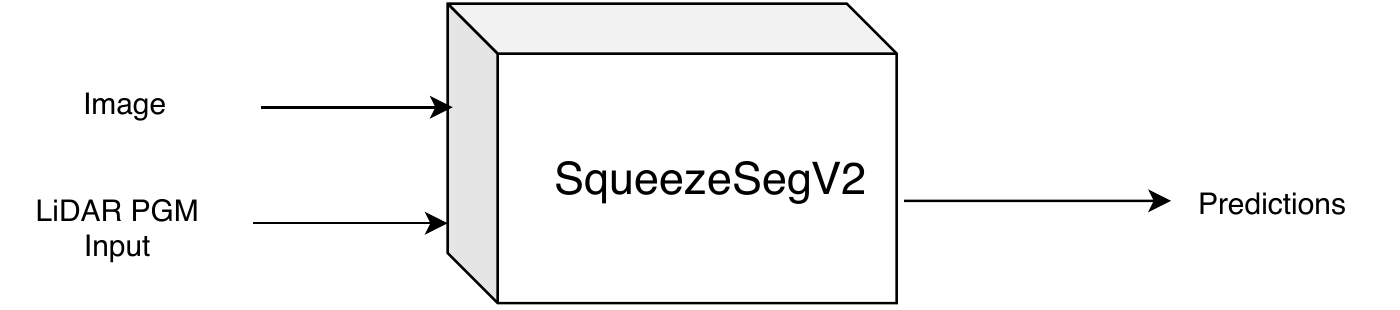}
  \caption{Early Fusion Model}
  \label{fig:early_fusion}
\end{figure}

\subsubsection{Mid-level Fusion}
\begin{figure}[htb]
    \centering
    \includegraphics[width=\linewidth]{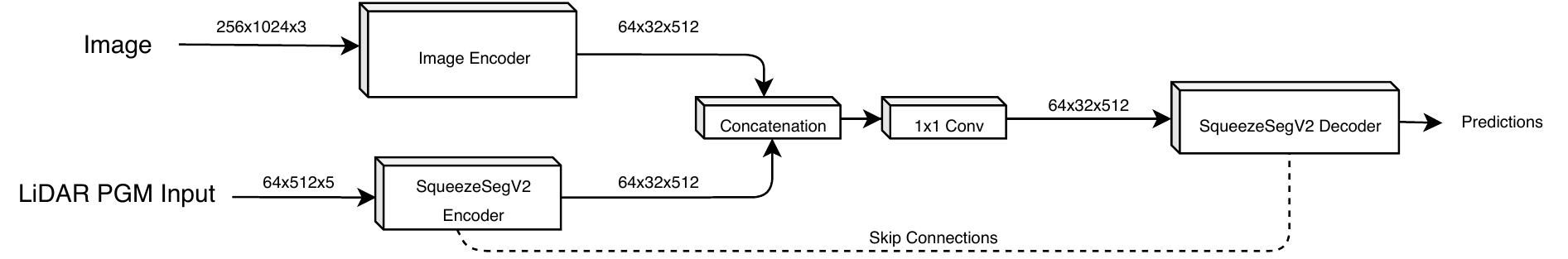}
    \caption{Mid-level Fusion}
    \label{fig:Mid_levl}
\end{figure}

\begin{figure*}[htb]
    \centering
    \includegraphics[width=\linewidth]{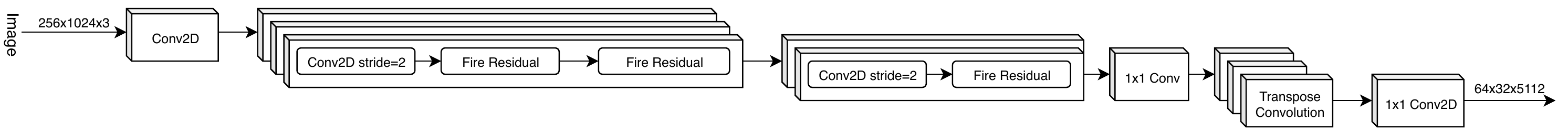}
    \caption{Image Encoder for Mid-level Fusion}
    \label{fig:ImageEncoder_Mid_levl}
\end{figure*}

Instead of combining information from the two modalities in the input space as in the early-level fusion, mid-level fusion combines information in an encoded feature space. The mid-level fusion model takes as input both the \(RGB\) image and the LiDAR PGM tensor with 5 channels representing \( x, y, z, intensity, range\). The input LiDAR tensor is fed to SqueezeSegV2 encoder while the image is fed to another fully convolutional encoder. The image encoder shown in figure \ref{fig:ImageEncoder_Mid_levl} uses a building block called fire residual which is the fire module with slight modification of adding a residual connection between its input and output. The outputs from the image and LiDAR encoders are concatenated together and passed down to additional fire modules. Finally, the output of these modules is fed to SqueezeSegV2 decoder to generate the predicated label map. The architecture of mid-level fusion model is shown in figure \ref{fig:Mid_levl}

\subsubsection{Late-level Fusion}
\begin{figure}[htb]
    \centering
    \includegraphics[width=\linewidth]{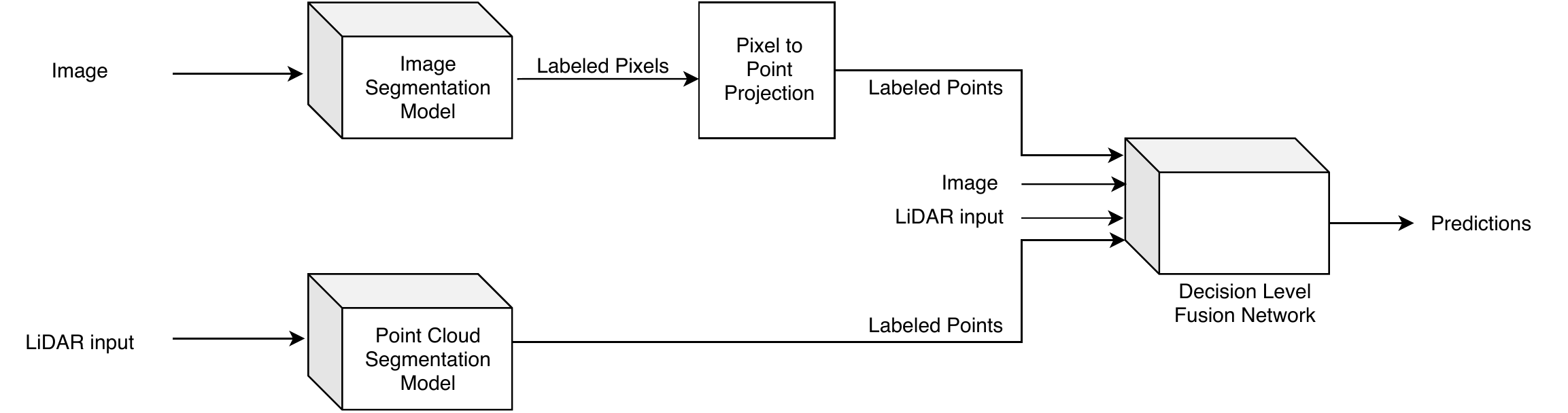}
    \caption{Decision Level Fusion Network}
    \label{fig:late_fusion}
\end{figure}

\begin{figure}[htbp]
    \centering
    \includegraphics[width=\linewidth]{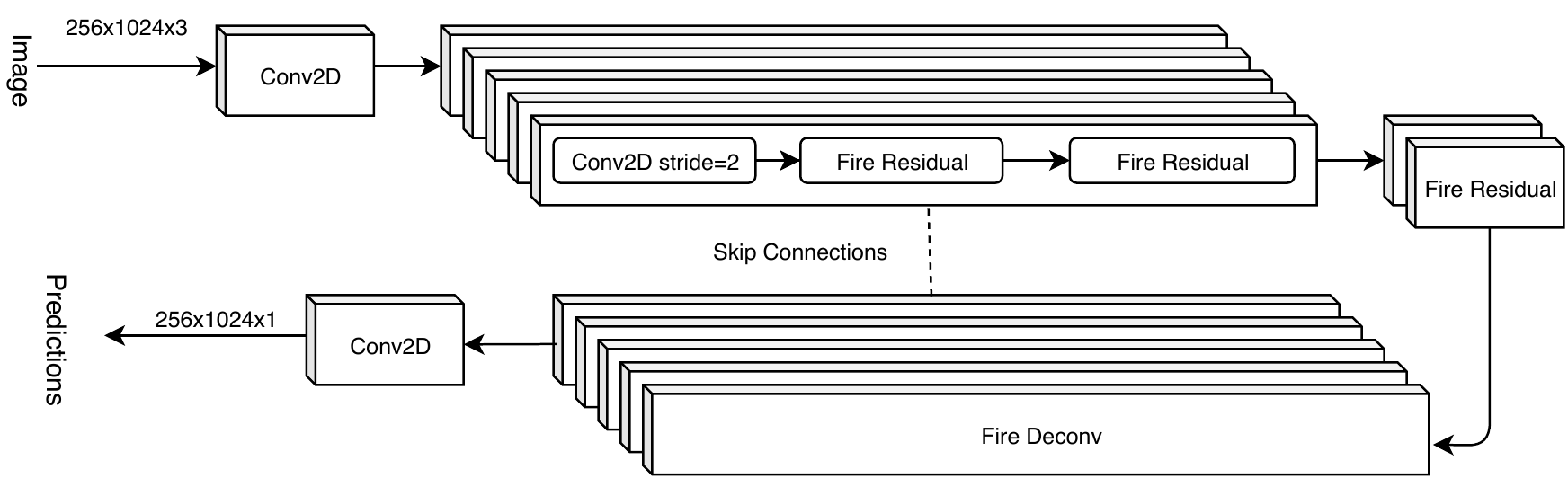}
    \caption{Image segmentation model for late fusion}
    \label{fig:image_seg_late_fusion}
\end{figure}

In an attempt to build upon the results of previous models, late-level fusion works by combining the outputs of two models that each work on either pixel or LiDAR space. An image segmentation model processes the \(RGB\) input data and produces a segmentation map of the images. Concurrently, a LiDAR semantic segmentation model is used on the \(XYZ\) data and produces a segmentation map of the point cloud. To combine both modalities, we have opted to work in the LiDAR space and project the pixels to their corresponding points using the extrinsic and intrinsic calibration matrices of the dataset. The resulting projection parameters are used to augment the \(x, y, z, intensity, range\) values with \(r, g, b\) ones in addition to two channels \(l1\) and \(l2\) that represent the output label map of both the image segmentation model and the point cloud segmentation model respectively. Proposed architecture is shown in figure \ref{fig:late_fusion}. The Early fusion model is used as point cloud segmentation model. The architecture for image segmentation model is shown in figure \ref{fig:image_seg_late_fusion}

\section{Experiments}
\label{section:sec_four}

\subsection{Dataset}
SemanticKITTI \cite{SemnaticKitti} is used to evaluate our proposed models. SemanticKITTI is composed of 21 sequences that were divided into training, validation and test set. The training set consists of sequences [00-06, 09-10] with a total of \(18,029\) scans. For validation, we use sequence [07] consisting of \(1,101\) scans and sequence [08] for testing which consists of \(4,071\) scans.
While SemanticKitti \cite{SemnaticKitti} provides semantic annotations for the LiDAR scans, no such annotations are provided for the images. To be able to train the late-fusion model which requires \(r,g,b\), we trained our image segmentation model on the popularly available CityScapes dataset \cite{cityscapes} and remapped both its labels and the SemanticKitti labels to a reduced merged version of them. Mappings from each respective dataset labels to our reduced version are provided in table~\ref{tab:class_mappings}

\begin{table}
    \caption{Class Mapping from SemanticKitti \cite{SemnaticKitti} and CityScapes \cite{cityscapes} datasets to our reduced version.}
    \label{tab:class_mappings}
    \centering
    \begin{tabular}{llll} 
    \toprule
    \multicolumn{4}{c}{Class Mappings} \\
    \cmidrule(r){1-4} 
    \multicolumn{2}{c}{SemanticKITTI} & \multicolumn{2}{c}{CityScapes}\\
    \cmidrule(r){1-2} \cmidrule(r){3-4}
    Original & Mapped & Original & Mapped \\
    \cmidrule(r){1-2} \cmidrule(r){3-4}
    bicyclist & rider & sky & unlabeled \\
    motorcyclist & rider & train & other-vehicle \\
    trunk & vegetation & wall & building \\
    parking & unlabeled & traffic-light & pole \\
    other-ground & unlabeld & - & - \\
    \bottomrule
    \end{tabular}
\end{table}

\newcommand*\rot{\rotatebox{90}}

\begin{table}
    \caption{SemanticKitti dataset results (15 classes) for all models on sequence 08 (test set). All methods were trained on sequences 00 to 10, except for sequence 07 which is used as validation set.}
    \label{tab:table1}
    \centering
    \resizebox{\textwidth}{!}{\begin{tabular}{llllllllllllllllll}
    \toprule
    Approach    & mIoU & OA & \rot{car} & \rot{bicycle}  & \rot{motorcycle} & \rot{truck} & \rot{other-vehicle} & \rot{person} & \rot{rider} & \rot{road} & \rot{sidewalk} & \rot{building} & \rot{fence} & \rot{vegetation} & \rot{terrain} & \rot{pole} & \rot{traffic-sign}  \\
    \cmidrule(r){1-1} \cmidrule(r){2-3} \cmidrule(r){4-18}
    LiDAR         & 0.411 & 0.844 & 0.827 & \textbf{0.096} & 0.170 & 0.049 & 0.176 & 0.117 & 0.448 & 0.943 & 0.735 & 0.658 & 0.231 & 0.713 & 0.681 & 0.107 & 0.216\\
    Early Fusion  & 0.419 & 0.858 & 0.836 & 0.052 & 0.148 & 0.067 & 0.206 & 0.117 & 0.391 & 0.939 & 0.731 & 0.685 & 0.260 & 0.765 & 0.676 & 0.118 & \textbf{0.293}\\
    Late Fusion   & 0.410 & 0.854 & 0.830 & 0.041 & 0.129 & 0.047 & 0.132 & 0.089 & 0.442 & 0.937 & 0.723 & 0.686 & 0.284 & 0.750 & 0.685 & 0.129 & 0.229 \\
    Late Fusion+D   & 0.427 & 0.865 & 0.827 & 0.062 & 0.139 & 0.074 & 0.130 & \textbf{0.173} & 0.452 & 0.937 & 0.724 & \textbf{0.723} & \textbf{0.316} & \textbf{0.783} & \textbf{0.686} & \textbf{0.143} & 0.242\\
    Mid Fusion    & \textbf{0.438} & \textbf{0.859} & \textbf{0.848} & 0.061 & \textbf{0.174} & \textbf{0.098} & \textbf{0.259} & 0.128 & \textbf{0.456} & \textbf{0.945} & \textbf{0.744} & 0.708 & 0.311 & 0.747 & 0.684 & 0.126 & 0.287\\
    \bottomrule
    \end{tabular}}
\end{table}

\subsection{Training}
For all experiments, weighted cross entropy loss is calculated according to equation \eqref{eq1} where \(C\) is the number of classes and \(f_c\) is the frequency of class \(c\) in the original SemanticKITTI dataset \cite{milioto2019iros}. Loss is optimized using stochastic gradient descent with initial learning rate \(0.01\) and momentum \(0.9\). Models are trained for at most \textbf{350} epochs with a batch size of \textbf{64} except in the middle level fusion model where batch size is reduced to \textbf{32} due to GPU memory constraints. Model evaluation is done on test set for best preforming weights on validation set. The performance metric used is class wise mean intersection over union given by \eqref{eq2} where $TP_c,FP_c,FN_c$ are the number of true positives, false positives and false negatives for class \(c\) respectively \cite{SemnaticKitti}. The unlabelled class is not included in loss calculations or performance metric calculations. We have built upon the code base provided by \cite{milioto2019iros, SemnaticKitti} for implementation of the SqueezeSegV2 model and for running the experiments. 

\begin{gather} \label{eq1}
L = -\sum_{c=1}^{C} w_{c}*y_{c}*\log({\hat{y_c}}) \\
\text{where}~\nonumber w_c = \frac{1}{log(f_c + \epsilon)}
\end{gather}

\begin{equation} \label{eq2}
\frac{1}{C} \sum_{c=1}^{C} \frac{TP_c}{TP_c + FP_c + FN_c}   
\end{equation}

\subsection{Discussion}
The proposed models were evaluated on the chosen test set of SemanticKitti. The results are shown in Table~\ref{tab:table1}. Table~\ref{tab:table3} shows inference times and number of parameters for our proposed models where Table \ref{tab:quantization_table} shows inference times, size, performance of our proposed model after quantization

\paragraph{Early-level fusion} Analysis of results obtained by fusion models show the significant value of fusing color information and LiDAR information for semantic segmentation. This is further supported by the fact that the early-level fusion model performs better that the LiDAR only baseline on most classes.
This approach, while having a good relative performance as shown in Table \ref{tab:table1}, is still not improving much on the baseline model and raises the question of its added benefit.

\begin{table}
    \caption{Performance measurements on SemanticKITTI dataset. Inference times are forward propagation times measured on Intel i7 CPU and are averaged for sequence 08 (test set).}
    \label{tab:table3}
    \centering
    \begin{tabular}{lll} 
    \hline
        \toprule
        Model & Inference Time (ms) & Number of Parameters\\ 
        \midrule
        LiDAR         & 401\;$\pm$10 & 926433\\
        Late Fusion   & 2594\;$\pm$20 & 3297525\\
        Early Fusion  & 397\;$\pm$10 & 928353\\
        Mid Fusion    & 642\;$\pm$38 &  2374641\\
        \bottomrule
    \end{tabular}
\end{table}

\begin{table}
    \caption{Quantization results}
    \label{tab:quantization_table}
    \centering
    \begin{tabular}{lllll}
        \toprule
        Approach &  PC(ms) & Before(MBs) & After(MBs) & mIoU\\
        \midrule
        Early-fusion & 127\;$\pm$8 & 3.81 & 1.05 & 0.400  \\
        Mid-fusion & 131\;$\pm$15 & 9.66 & 2.68 & 0.101\\
        \bottomrule
    \end{tabular}
\end{table}

\paragraph{Midl-level fusion}
The obtained results show the mid-level fusion model as the best performing model. However, a closer analysis to individual class show more mixed results. For example, the model preformed better than Late fusion model on classes such as car, sidewalk and road while it preformed worse on classes such as bicycle and person. Table \ref{tab:table1} shows us that the mid-level fusion has approximately twice the number of parameters of early fusion model with about \textbf{2\%} improvement in performance. This may reduce the appeal of mid-level fusion model. However, the image encoder was not pretrained for this experiment. Having a pretrained image encoder might have boosted the model's performance especially for those classes that are less frequent in dataset.

\paragraph{Late-level fusion}
The late-level fusion model while improving upon the LiDAR only model, has achieved lower results than the mid-level fusion method as outlined in Table \ref{tab:table1}. This can be mostly attributed to the fact that the model draws on the previous decisions achieved by the two image and LiDAR segmentation models and is affected by their shortcomings and biased to them in nature. While it can achieve a modicum of improvement, it is still limited by the previous models performance. 
With regards to performance, Table \ref{tab:table3} shows the late fusion to hold the worst inference time across all models. This is mostly due to that the model consists of three separate models and inference time for all three was taken into account with the image segmentation model taking the biggest chunk of the inference time. Late Fusion results are reported for two approaches with the +D approach using deepLab as its image segmentation model.
\paragraph{Quantization} Due to the tight memory and processing constraints of computing in autonomous vehicles, it was essential to reduce the inference time of our approaches as well as their memory footprint by using one of the quantization techniques available. For this paper we settled on Post-Training Quantization which did not require re-training the network and was easily done by inserting observers to understand the range of values coming out of the different activation layers in the network. The technique changes weights used from floating point to INT8 weights which in our models resulted in about \textbf{3.6} times reduction in size and faster inference time. Table \ref{tab:quantization_table} shows the results obtained in early and mid-level fusion as these were the best performing models with the former being the fastest and the latter, more accurate. For the mid-fusion, the result of quantization is drastically worse and requires further investigation.

\section{Conclusions}
\label{section:sec_five}
In this paper, we explored the added benefit of fusing input from two different sensors for the task of semantic segmentation within the context of autonomous driving. By exploring the results of non-fusion approaches and the different types of fusion, we aim to show that a fusion model outperforms the baselines of same comparable architecture. Towards this end, we run models based on the popular SqueezeSegV2 \cite{squeezesegv2} model that utilizes three different fusion levels and show that fusion outperforms the baseline methods in all levels with the mid-level fusion achieving the most improvement.

\section{Future Work}
\label{section:sec_six}
Our work shows that fusion approaches improve upon existing models. However, such improvements are still low in percentage and require further exploration to enhance. It is also imperative to explore the concepts introduced in this paper with different models for further comparison.

\begin{ack}
The authors would like to thank Valeo Internal Automotive Software Egypt (LLC) for sponsoring this project.
\end{ack}


\bibliographystyle{plainnat}
\bibliography{bibliography}

\end{document}